\documentclass[10pt]{article}

\usepackage{float}

\newif\ifdegree
\degreetrue

\relax

\usepackage{graphicx}
\usepackage{amssymb}
\usepackage{amsmath}
\usepackage{wrapfig}
\usepackage{url}

\ifdegree

\fi

\DeclareMathOperator*{\argmax}{arg\,max}

\newcommand{\bff}{\mathbf{f}}
\newcommand{\bfr}{\bff^{R}}

\newcommand{\bfz}{\mathbf{z}}

\newcommand{\bs}[1]{\boldsymbol #1}

\newcommand{\bmu}{\bs\mu}

\newcommand{\bSigma}{\bs\Sigma}

\newcommand{\phuman}{p(\bfh\mid\bfz^{h}_{1:t})}

\newcommand{\bfh}{\mathbf{h}}

\newcommand{\fa}{p(\bfh, \bff^{R},\bff\mid\bfz_{1:t})}

\usepackage{fullpage}
\usepackage{geometry}
\usepackage{fancyhdr}
\usepackage{graphicx}
\usepackage{wrapfig}
\usepackage{url}

\renewcommand{\title}[2]{\def\title{#1}\def\shorttitle{#2}}
\renewcommand{\author}[2]{\def\author{#1}\def\shortauthor{#2}}

\renewcommand{\maketitle}{
  \thispagestyle{plain}
  \section*{\title}
  \subsubsection*{\author}
  \vspace{1em}
  \setcounter{page}{1}
}

\usepackage{wrapfig}
\usepackage{subfig}  
\usepackage{hyperref}
\usepackage[square,numbers,sort&compress]{natbib}
\usepackage[font={footnotesize}]{caption}
        
\usepackage{epsfig}
\usepackage{amsmath,amssymb,dsfont,amsthm}
\usepackage{tikz}                                                        
\usepackage{graphicx}
\usepackage{amssymb}
\usepackage{epstopdf}
\usepackage{graphics}
\usepackage{amsmath, amssymb}
\numberwithin{equation}{section}
\usepackage{verbatim}   
\usepackage{color}      
\usepackage{subfig}  
\usepackage{hyperref}   
\usepackage{xspace}
\usepackage{float}
\usepackage[font={footnotesize}]{caption}
\usepackage{wrapfig}
\usepackage{natbib}

\title{A Mathematical Theory of Human Machine Teaming}{A Mathematical Theory of HMT}
\author{Pete Trautman}
       {}

\begin{document}

\maketitle

\begin{wrapfigure}{r}{0.45\textwidth}
\centering 
\includegraphics[width=0.45\textwidth]{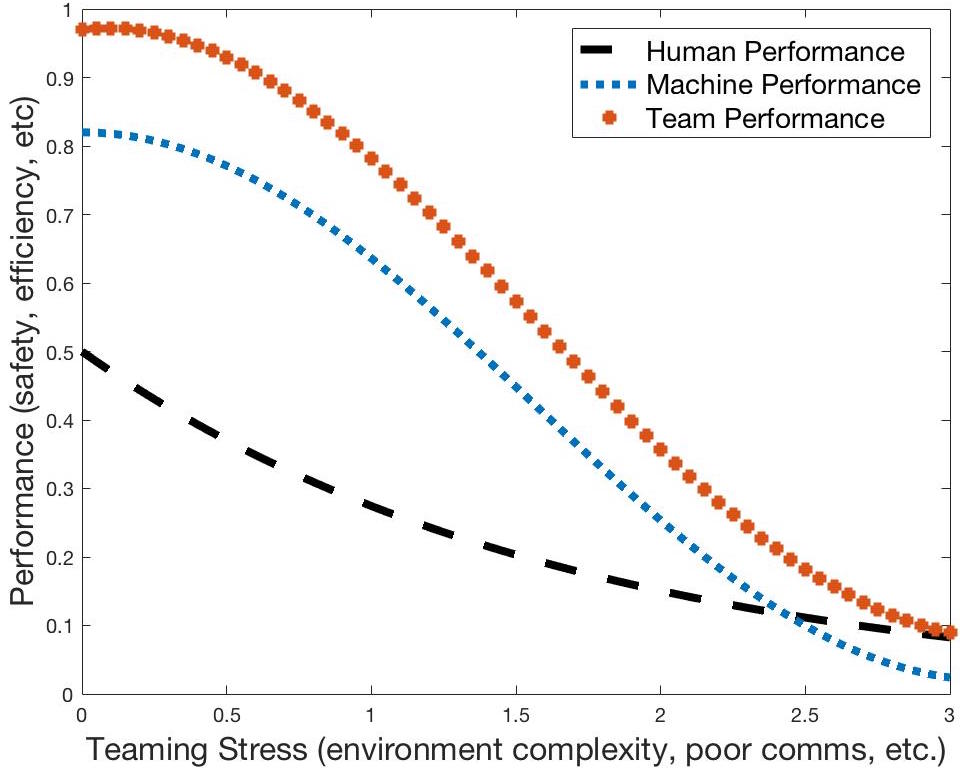}
\vspace{-0pt} 
\caption{Illustration: an HMT architecture that is lower bounded for generic performance metrics under a variety of teaming stressors.}
\label{fig:lower-bound}
\vspace{-0pt}
\end{wrapfigure}
We begin with a disquieting paradox: human machine teaming (HMT) often produces results worse than either the human or machine would produce alone. Critically, this failure is not a result of inferior human modeling or a suboptimal autonomy: even with perfect knowledge of human intention and perfect autonomy performance, prevailing teaming architectures still fail under trivial stressors~\cite{trautman-smc-2015}.  This failure is instead a result of deficiencies at the \emph{decision fusion level}.  Accordingly, \emph{efforts aimed solely at improving human prediction or improving autonomous performance will not produce acceptable HMTs: we can no longer model humans, machines and adversaries as distinct entities.}   We thus argue for a strong but essential condition: HMTs should perform no worse than either member of the team alone, and this performance bound should be independent of environment complexity, human-machine interfacing, accuracy of the human model, or reliability of autonomy or human decision making.  In other words, this requirement is  \emph{fundamental} (Figure~\ref{fig:lower-bound}): the fusion of two decision makers should be as good as either in isolation.  For instance, if the human model is incorrect, the performance of the team should still be as good as the autonomy in isolation.  If the autonomy is unreliable, this should not impair the human.  Importantly, most existing HMTs do not have a robust mechanism to ``fuse'' human and machine information, thus obviating any opportunity at producing ``a team that performs greater than the sum of its parts''.  In response to these shortcomings, we introduce a theory of \emph{interacting random trajectories} (IRT) over the humans, machines, and (potentially adversarial) environment~\cite{trautman-smc-2015} that optimally fuses the three sources of information and achieves the following four objectives:
\vspace{-0pt}
\begin{enumerate}
\item IRT is a \emph{unifying} formalism: most HMT approaches are  approximations to IRT.
\vspace{-0pt}
\item IRT \emph{quantifies} these approximations, and precisely predicts when architectures will fail.
\vspace{-0pt}
\item We can \emph{predict}, in advance of empirical evaluation, when IRT will succeed and fail.
\vspace{-0pt}
\item The first three objectives, when combined with dimensionality reduction techniques, enable a human-collective/multi-agent decision fusion framework with performance bounds.
\end{enumerate}

\vspace{-0pt}
\paragraph{1.  A Unifying Formalism for HMT}
\begin{wrapfigure}{r}{0.55\textwidth}
\centering \vspace{-17pt}
\includegraphics[width=0.55\textwidth]{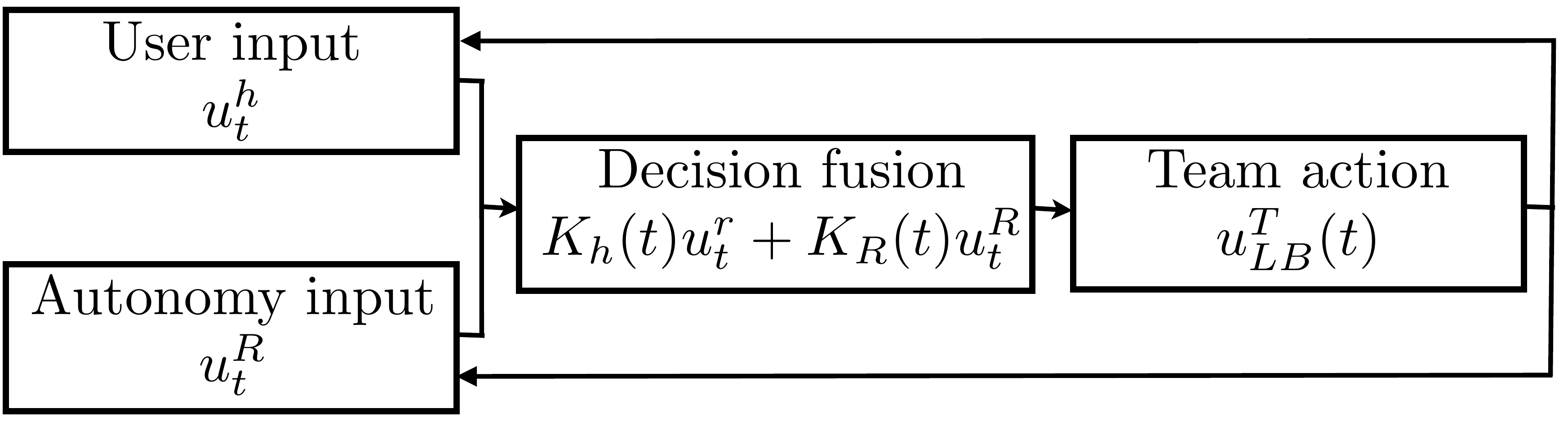} 
\vspace{-20pt}
\caption{State of the art HMT architectures (shared control, task allocation, autopilots, and HCI) are not reliably lower bounded. ``Input'' ranges from high to low level.}
\label{fig:lin-blend}
\vspace{-0pt}
\end{wrapfigure}
To show that IRT is a unifying formalism, we must understand standard HMT decision fusion; we thus introduce linear blending:
\begin{align}
\label{eq:lin-blend}
u^T_{LB}(t) = K_h(t)u^h_t + K_R(t)u^R_{t}.
\end{align}
At time $t$, $u^T_{LB}(t)$ is the team action, $u^h_t$ is the human operator input (joystick deflections, high level commands, or preset autonomous actions), $u^R_{t}$ is the autonomy command, and $K_h(t), K_R(t)$ are the operator and autonomy weighting factors, respectively, which can be functions of anything, subject to $K_h+K_R=1$.  As shown in~\cite{trautman-smc-2015}, linear blending captures a wide variety of teaming approaches in \emph{low level} shared control---we argue here that linear decision fusion is used much more broadly in HMT.  For instance, switching control (either human or machine has full control) is a special case of linear blending where $K_h, K_R$ are either 0 or 1. Consider the following: 
\vspace{-0pt}
\begin{itemize}
\item Dynamic task allocation: an algorithm determines when the human or the machine should be in control of the task (switching control).  This does not mean that the \emph{allocation method is linear}, but that the \emph{decision fusion is linear}.   For instance, we might use an elaborate cognitive architecture to determine when the human takes full control.  We point out that ``shared mental models'' are typically implemented in a switching control fashion, supervisory control approaches delegate to either the machines or the humans, and handoff tasks have been almost exclusively restricted to the case of switching control.
\vspace{-0pt}
\item Commercial autopilots are exclusively switching control.
\vspace{-0pt}
\item ``Playbook'' approaches: the human picks a ``play'', and the machines execute, which is an example of switching control where $K_h(0)=1,K_R(1)=1$.
\vspace{-0pt}
\item Standard HCI: human inputs data, machine processes data/presents alternatives to human, human makes a decision.  This is switching control: $K_h(0)=1,K_R(1)=1,K_h(2)=1$.
\end{itemize}
\vspace{-0pt}
\emph{Interacting random trajectories} \cite{trautman-smc-2015} is a generalization of linear blending: \textbf{it is a statistically sound and optimal approach to fusing coevolving human, machine and environment information}.  IRT relaxes human \emph{input} to \emph{online data} $u^h_t \doteq \bfz^h_{t}$ about the random human trajectory $\bfh \colon t\in\mathbb R \to \mathcal X^h$ over the action space $\mathcal X^h$ \emph{(we generalize to multiple humans in Section 4)}, we take measurements $\bfz^{R_j}_{1:t}$ of the $j$'th machine trajectories $\bff^{R_j}\colon t\in\mathbb R \to \mathcal X^R$, and measurements $\bfz^{f_i}_{1:t}$ of the $i$'th environment agent trajectory $\bff^i\colon t\in\mathbb R \to \mathcal X^f$.  We collapse the machines $\bfr = [\bff^{R_1},\ldots,\bff^{R_{m_t}}]$ and the environment $\bff = [\bff^{1},\ldots,\bff^{n_t}]$ into collective random processes, and take the following as our \emph{decision fusion architecture} (Equation~\ref{eq:irt} is updated at every $t$, so reflects \emph{collective evolution}):
\vspace{-0pt}
\begin{align}
\label{eq:irt}
\vspace{-0pt}
u^T_{IRT}(t) &= \bff^{R^*}_{t+1}\nonumber \\
 (\bfh,\bff^{R},\bff)^* &=\argmax_{\bfh, \bfr,\bff} p(\bfh, \bfr,\bff \mid \bfz^h_{1:t}, \bfz^R_{1:t},\bfz^f_{1:t}).
 \vspace{-0pt}
\end{align}
This formulation enables a precise understanding of the assumptions that \emph{any} linear fusion architecture imposes on the team, thus providing a theoretical advantage, since we can now do analysis \emph{in advance of empirical evaluation} across a broad range of approaches.  
 \vspace{-0pt}
\begin{figure}[t!]
\label{fig:cafe}
\centering
\subfloat[]{
\includegraphics[width=0.32\textwidth]{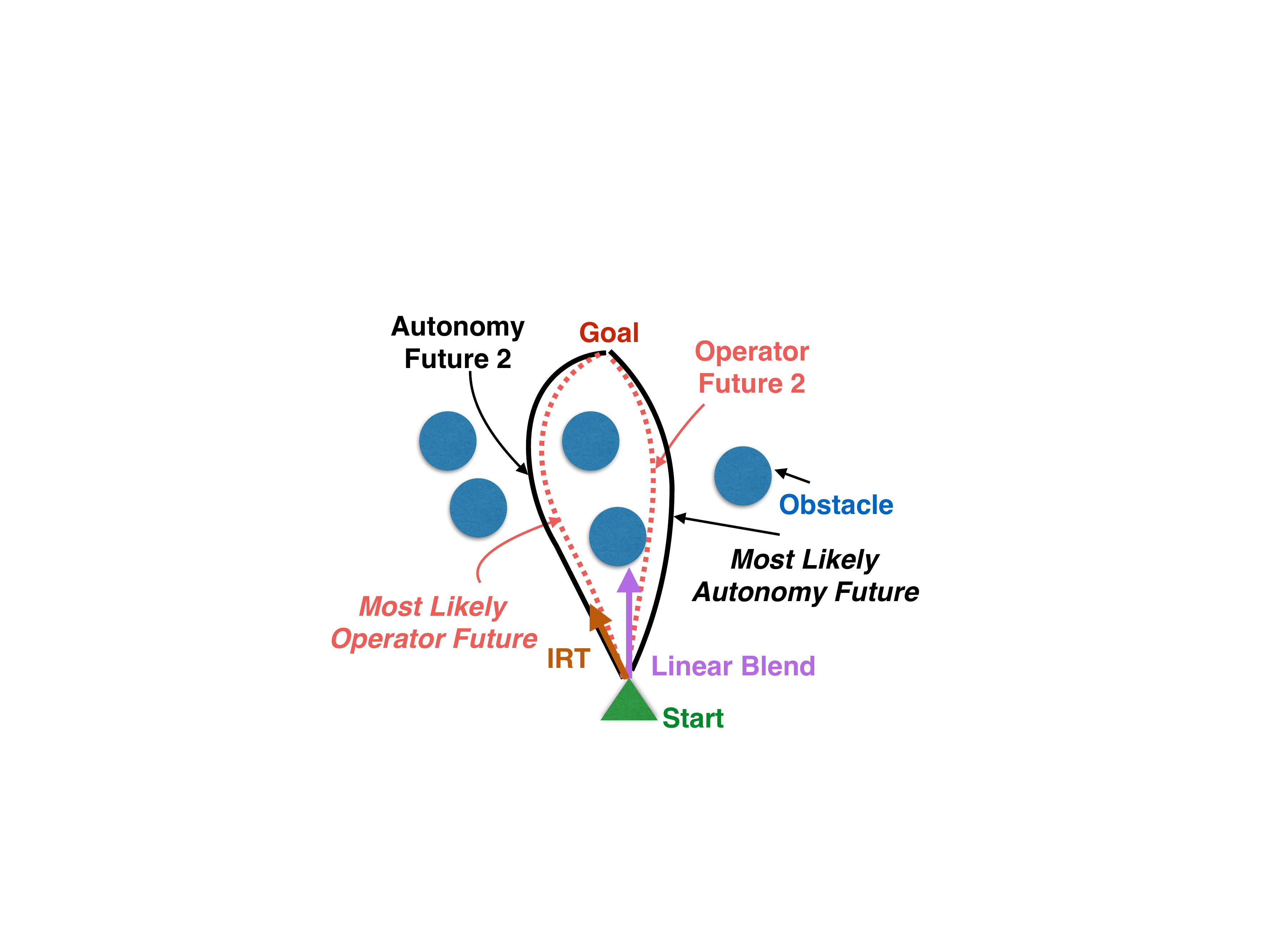}
\label{fig:linear-blend-fail}
}
 \subfloat[]{
    \includegraphics[width=0.32\textwidth]{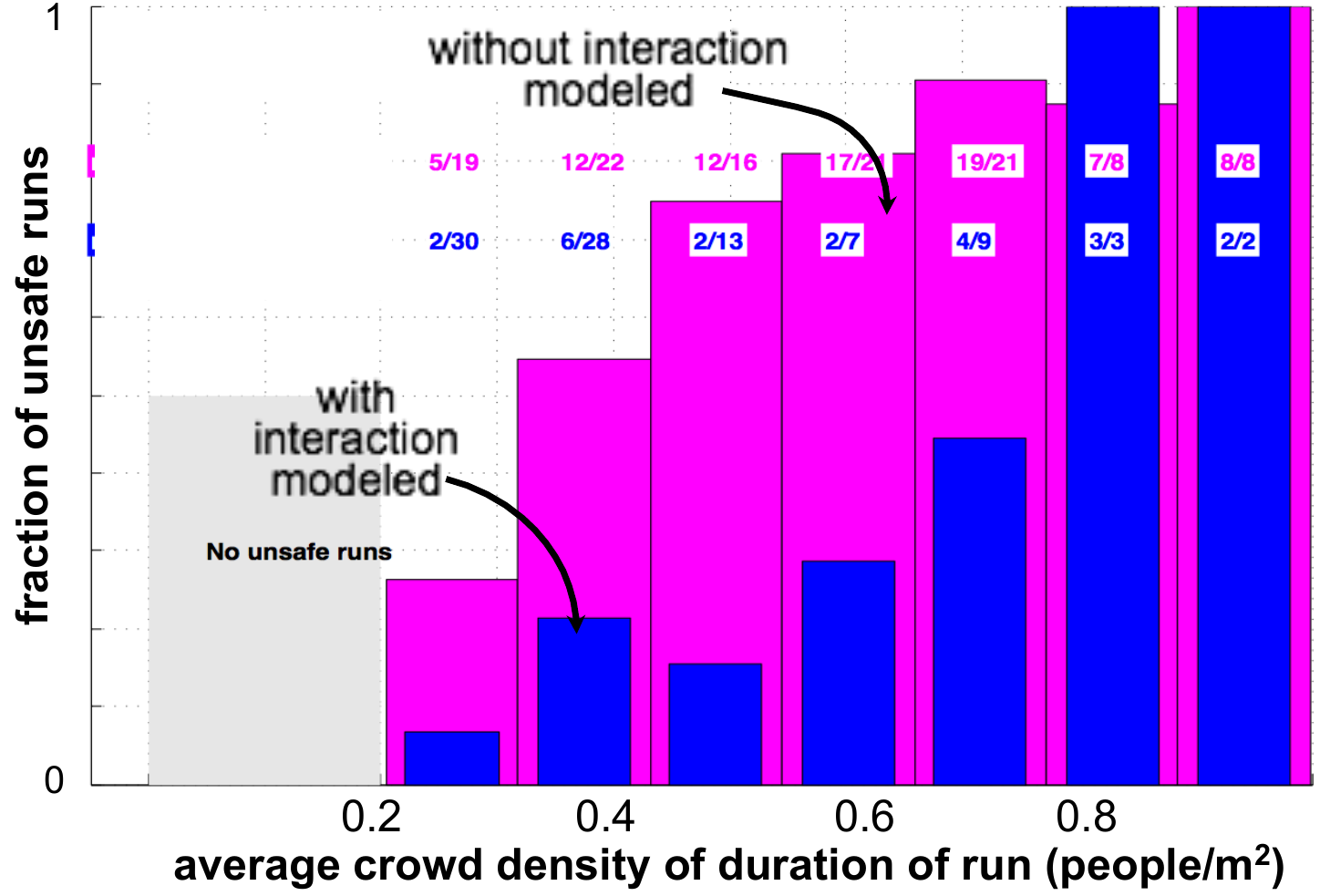}
    \label{fig:cooperative_advantage}
  }
\subfloat[]{
\includegraphics[width=0.32\textwidth]{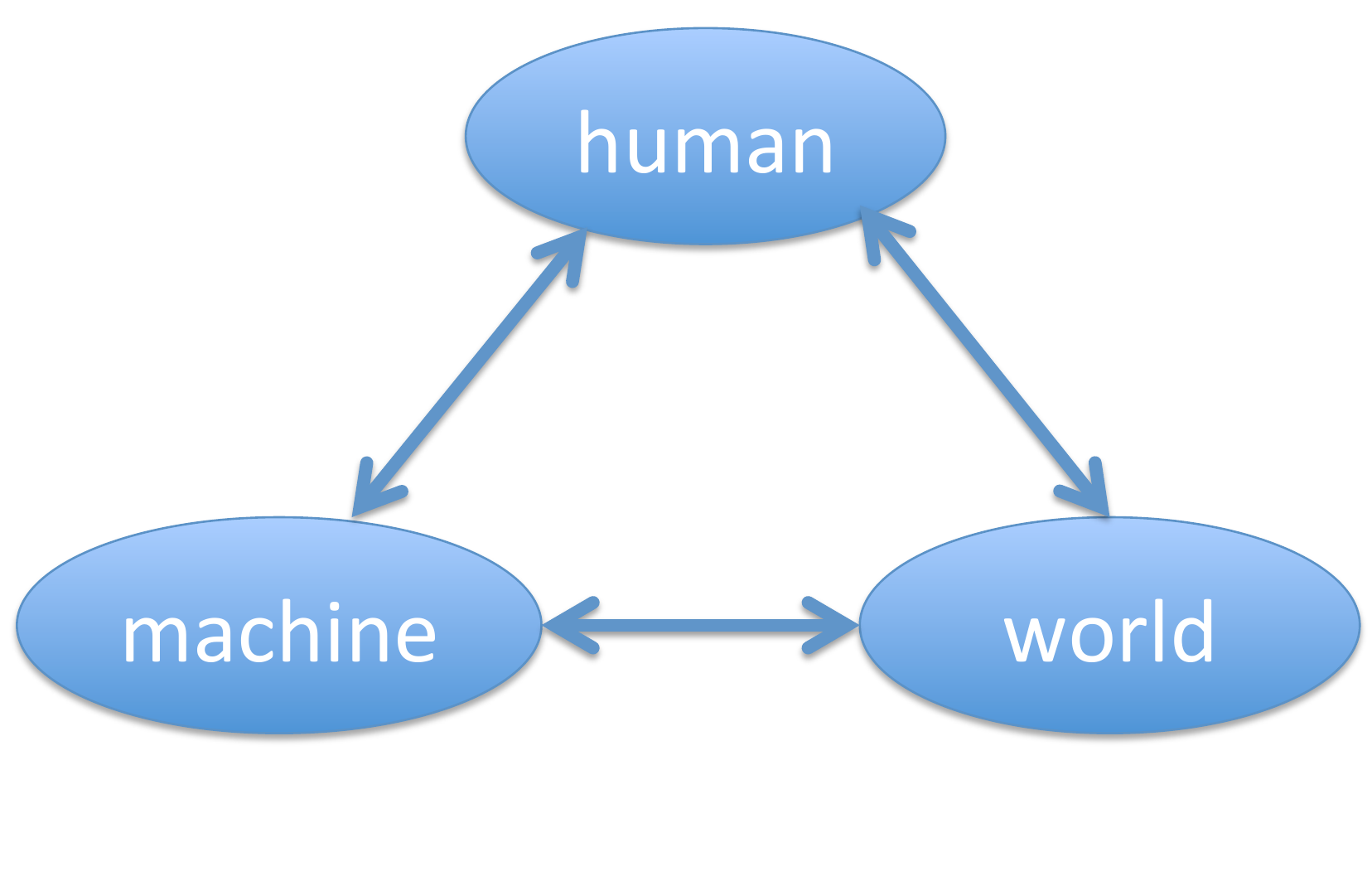}
\label{fig:triangle}
}
\vspace{-4mm}
\caption{\footnotesize{\textbf{(a)} Operator goes left and autonomy goes right, linear architectures fail lower bounding, IRT preserves lower bounding.  \textbf{(b)} Coupled robot-crowd models improves safety 3-fold, maintain efficiency near that of a human operator.  Decoupled models fail lower bounding property: purely human crowds outperform human-robot crowds. \textbf{(c)} \cite{trautman-smc-2015,irt-discuss} suggest a coupled human-machine-world architecture is required to achieve the lower bounding property.  }}
\vspace{-0pt}
\label{fig:cafeteria_testbed}
\end{figure}

\begin{figure}[t!]
\centering
\subfloat[]{
\includegraphics[width=0.34\textwidth]{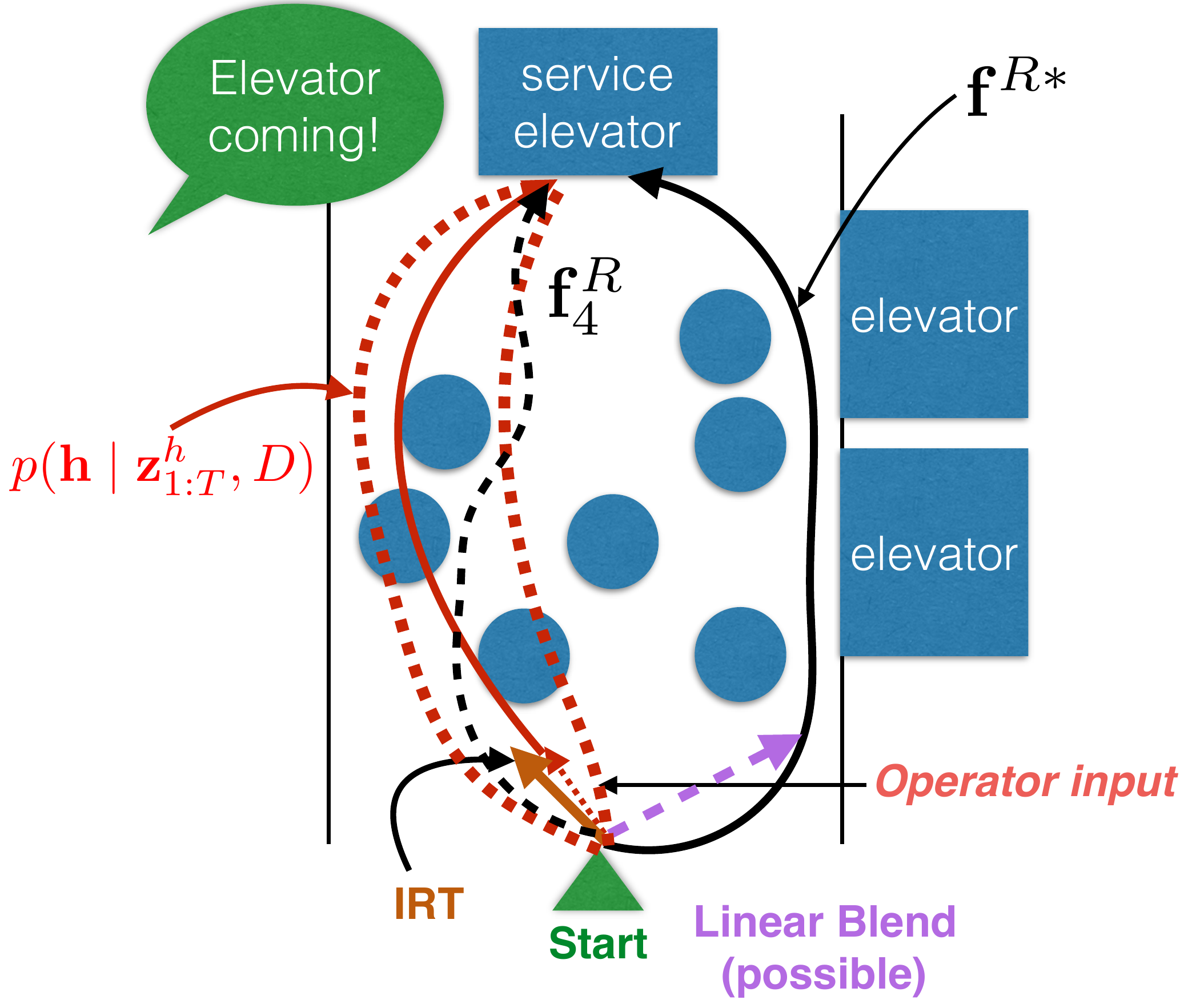}
\label{fig:hi-fi}
}
 \subfloat[]{
    \includegraphics[width=0.34\textwidth]{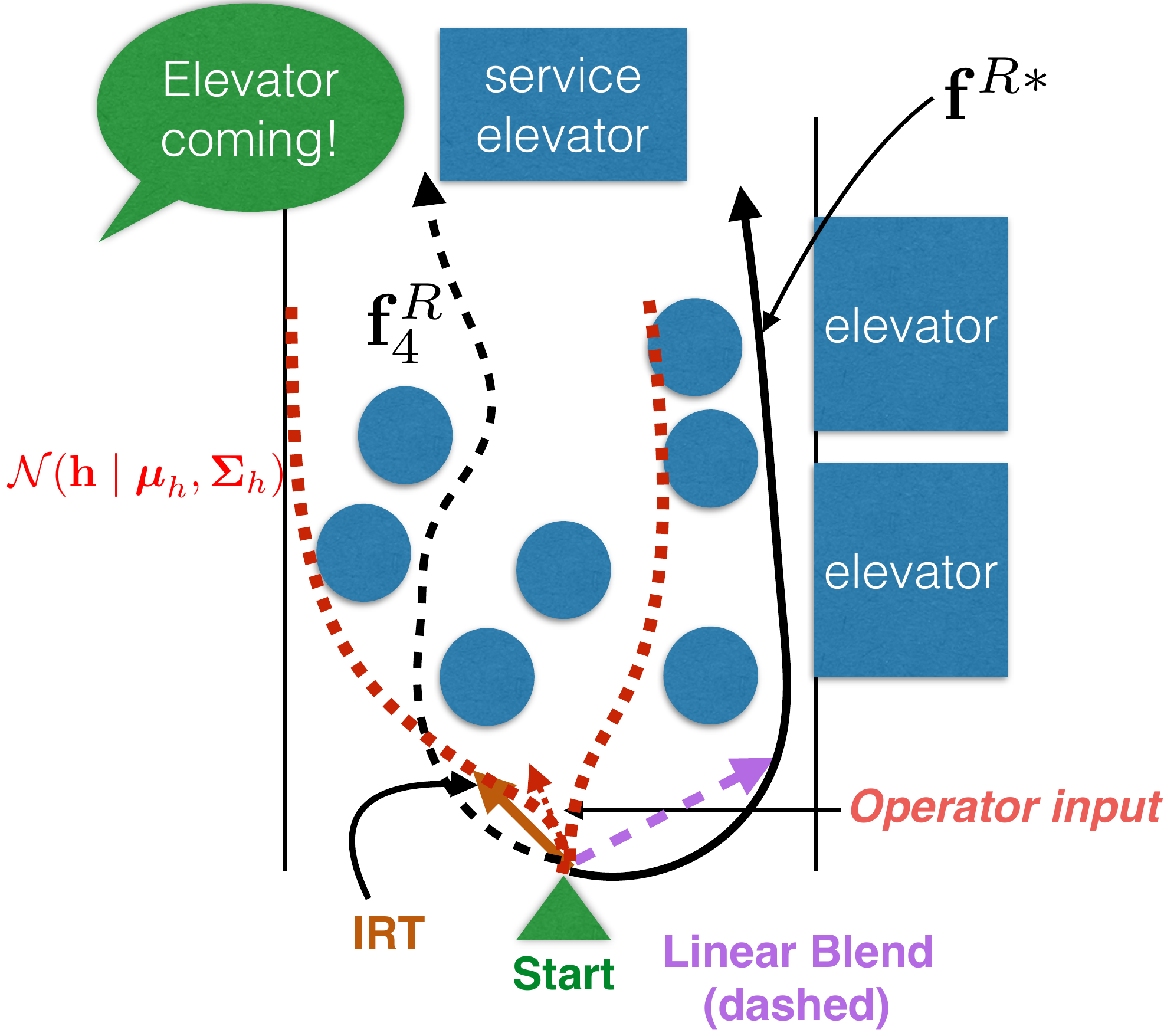}
    \label{fig:lo-fi}
  }
\subfloat[]{
\includegraphics[width=0.34\textwidth]{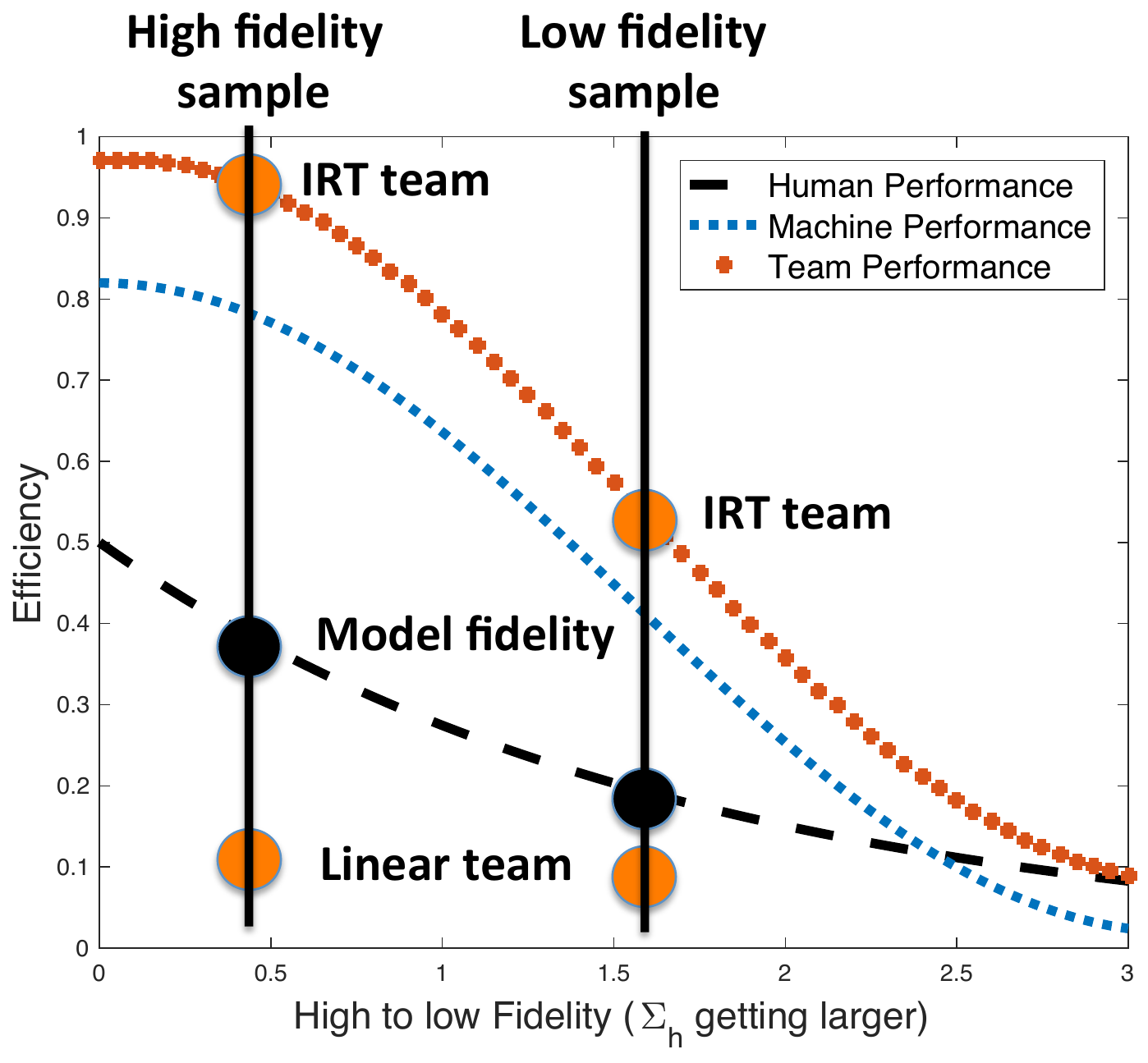}
\label{fig:fidelity-lower-bound}
}
\vspace{-0pt}
\caption{\footnotesize{Linear, IRT architectures navigating through crowd with semantic information (``elevator coming!'') \textbf{(a)} High fidelity human intent model ($p(\bfh \mid \bfz^h_{1:T},D)=\mathcal N(\bfh \mid \bmu_h,\bSigma_h)$ has small $\bSigma_h$): IRT able to leverage human contextual information; linear architectures violate lower bounding property.  \textbf{(b)} With large $\bSigma_h$, IRT team performance lower bound maintained.  Lower bound violated for linear approach.  \textbf{(c)} How to use data to instantiate Figure~\ref{fig:lower-bound}.  }}
\vspace{-0pt}
\label{fig:fidelity}
\end{figure}

\vspace{-0pt}
\paragraph{2.  Quantifying the Approximations in Existing HMT Approaches}
An important motivation for the lower bounding principle is the commonplace failure of existing HMT architectures---for many applications, trivial stressors lead to the team falling apart.   Consider the case of a human and a robot sharing control of a platform in a congested environment (e.g., a shared control wheelchair navigating through a crowd).  In~\cite{trautman-smc-2015}, we proved that state of the art HMT architectures fail the lower bounding criteria \textbf{if environmental or operator predictions are multimodal in a Gaussian mixture representation}; even under mildly challenging conditions, existing approaches can fuse two safe inputs into an \emph{unsafe} shared control (Figure~\ref{fig:linear-blend-fail}).  Furthermore, in~\cite{trautman-smc-2015}, we proved that IRT respects the lower bounding property under a variety of circumstances (explored fully in~\cite{irt-discuss}).  Most existing approaches to fully autonomous navigation in human environments also fail the lower bounding property. For instance, as shown in~\cite{trautmaniros, trautmanicra2013}, decoupling the components of the robot-crowd team leads to the \emph{freezing robot problem}: once environment complexity exceeds a certain threshold, planning algorithms that independently model the human and the robot freeze in place.  More broadly, as shown in Figure~\ref{fig:cooperative_advantage}, state of the art crowd navigation algorithms fail the lower bounding property: purely human crowds  are safer and more efficient than human-robot crowds. The results in~\cite{trautman-smc-2015,irt-discuss} suggest that a \emph{coupled and evolving} human-machine-world architecture is required to achieve the lower bounding property (Figure~\ref{fig:triangle}).  Further, linear fusion prevents the interleaving of human/machine capabilities, while IRT weaves complementary capabilities (Figure~\ref{fig:fidelity}).

\vspace{-0pt}
\vspace{-0pt}
\paragraph{3.  Predicting Capabilities of IRT} \begin{wrapfigure}{r}{0.35\textwidth}
\centering \vspace{-0pt}
\includegraphics[width=0.35\textwidth]{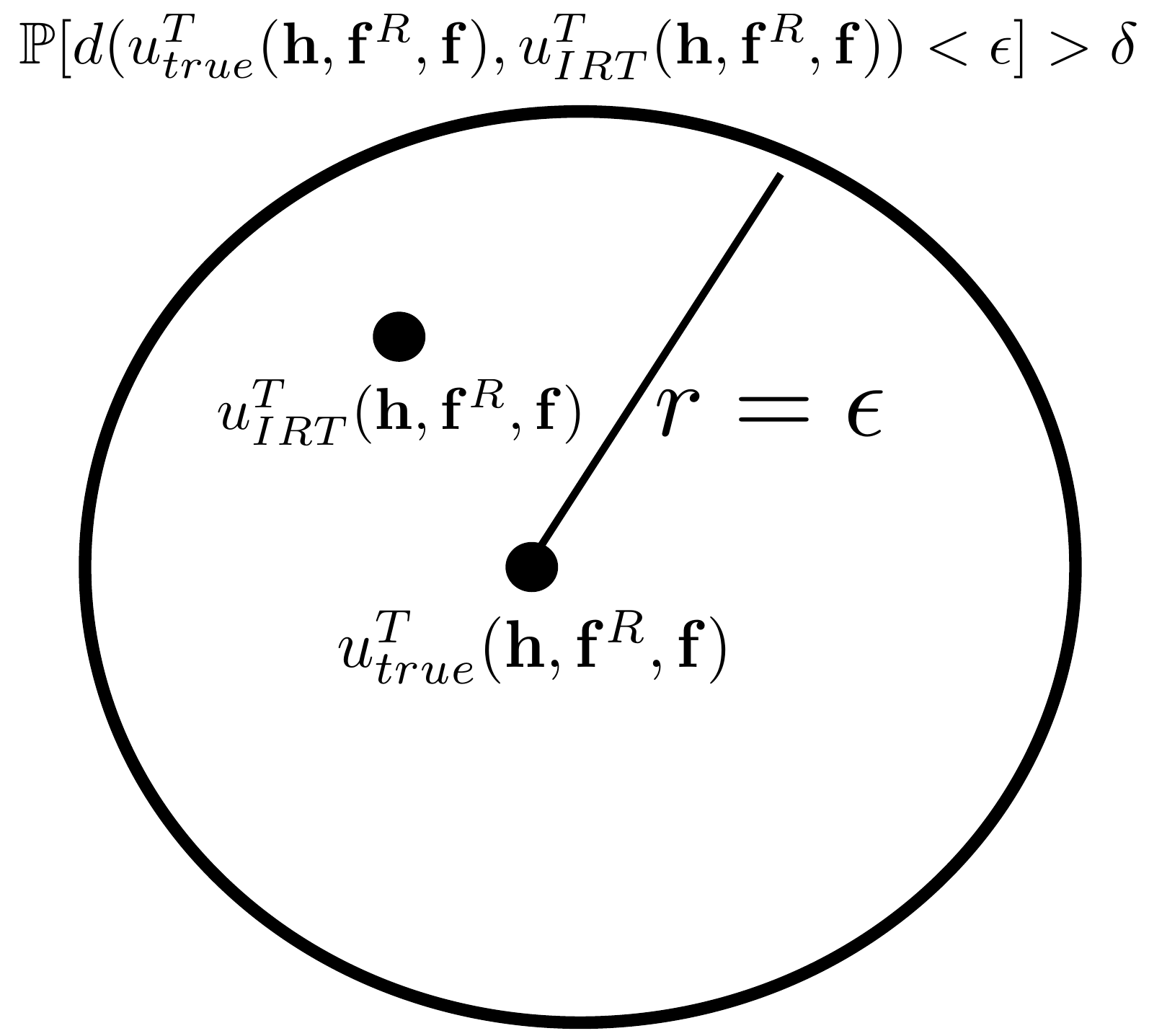} 
\vspace{-0pt}
\caption{Using learning theory and~\cite{trautman-smc-2015} to generate Figure~\ref{fig:lower-bound}. The ``true'' teaming action $u^T_{true}$ and IRT teaming action $u^T_{IRT}$ are separated by $\epsilon$ with probability $\delta$. The human, machine, and world ($\bfh,\bfr,\bff$) are arguments of the teaming actions; we vary them to produce Figure~\ref{fig:lower-bound}.}
\label{fig:pac-bound}
\vspace{-0pt}
\end{wrapfigure}Although a coupled HMT architecture is necessary to achieve lower bounding, can we \emph{prove} that joint models preserve the property across a spectrum of team stressors?  In Figure~\ref{fig:fidelity}, we present a thought experiment showing how we can reason towards Figure~\ref{fig:lower-bound}: a shared control robot travels through a crowd waiting for an elevator.  Without elevator arrival information, the robot's best choice is to go right.  When the elevator bell rings, the robot does not know what it means; a human, however, will recognize that the best path will be around the left side of the crowd and the \emph{worst} path will be to the right.  In Figure~\ref{fig:hi-fi}, a high fidelity human model shows how IRT marries human information to robot path planning to exceed the lower bound; the linear architecture discards the human's input and violates the lower bound (Figures~\ref{fig:hi-fi} and~\ref{fig:fidelity-lower-bound}).  In Figure~\ref{fig:lo-fi}, IRT exceeds the lower bound and the linear architecture violates it (Figures~\ref{fig:lo-fi},~\ref{fig:fidelity-lower-bound}).
\vspace{-0pt}

Constructing general instantiations of Figure~\ref{fig:lower-bound} will require at least two advances.  First, we must quantify \emph{performance} error outside of the training set.  IRT formulates teaming as a joint human-machine-world model, presenting an opportunity to interpret \emph{performance} error as \emph{generalization} error; such an approach allows us to leverage important results from learning theory, and thus accelerate our understanding of HMT performance bounds.  Second, to generate Figure~\ref{fig:lower-bound},  team stressors must be an argument of the predicted performance error.  Since the human, autonomy, and world are arguments of IRT, \emph{these models are also arguments of the performance error} (Figure~\ref{fig:pac-bound}). 

\vspace{-0pt}
\paragraph{4.  Optimal Human Collective Representations}
\label{sec:multiple-agents}
\begin{wrapfigure}{r}{0.45\textwidth}
\centering \vspace{-0pt}
\includegraphics[width=0.45\textwidth]{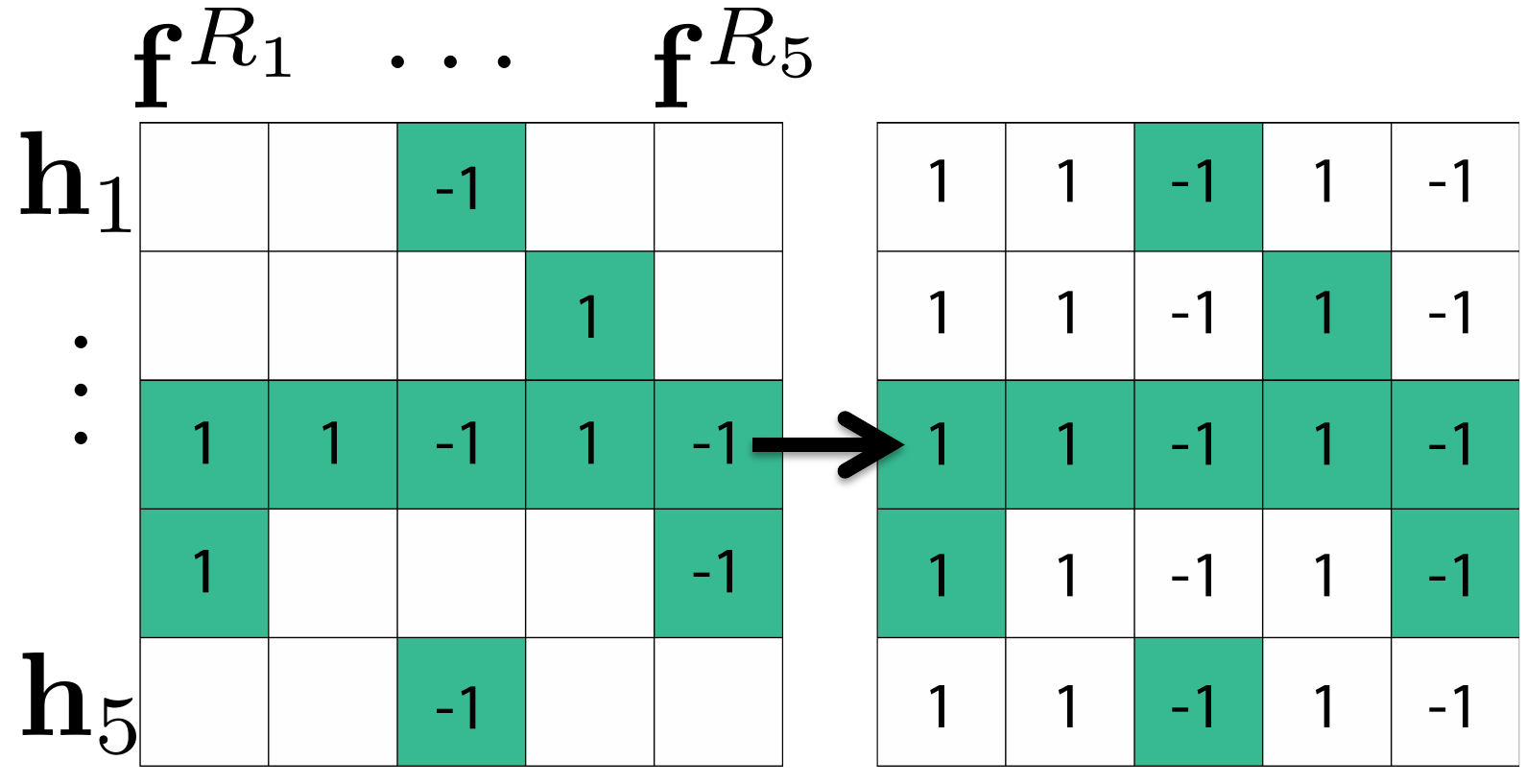} 
\vspace{-0pt}
\caption{From \url{https://en.wikipedia.org/wiki/Matrix_completion}: matrix completion of an $n_1\times N$ matrix, rank $r=1$.  The columns are $N$ ``machines'' and rows are $n_1$ ``operators''.}
\label{fig:matrix-complete}
\vspace{-0pt}
\end{wrapfigure}
 Many existing internet recommendation technologies (e.g., Netflix) are based on a simple observation: human preference is often low dimensional.  Thus, when represented in the optimal basis, we can accurately predict human decision making from only a few example decisions.  With matrix completion~\cite{matrixcompletion}, we can \emph{exactly} recover user preference for any movie with just a few reviews; \emph{however, this technology assumes that $n_1$ other humans have already entered $k$ reviews, blurring the distinction between individual and collective}.  This begs a critical question for HMT: can we infer \textbf{collective} human decision making from just a few individual operator samples?  We present three examples, and discuss the underlying mathematical challenge. 
\vspace{-0pt} 
\begin{enumerate}
\item Supervisory control of $N$ platforms: a single operator provides $n \ll N$ waypoint inputs.
\vspace{-0pt}
\item Human control of a robotic prosthesis: in a robotic prosthesis, there are $N$ actuators, but the human can only provide $n \ll N$ actuator inputs.
\vspace{-0pt}
\item Big data analysis (find the bad guy in 1B images): an analyst can provide up to $N$ image ``insights'' (derived from contextual clues).  With time pressure, he provides $n \ll N$ insights.
\end{enumerate}
\vspace{-0pt}
IRT, as described in Equation~\ref{eq:irt}, provides a \emph{mathematical quantification} of this problem:
\vspace{-0pt}
\begin{align*}
\fa = p(\bfr, \bff \mid \bfz^R_{1:t},\bff_{1:t},\bfh)\phuman 
\approx p(\bfr, \bff \mid \bfz^R_{1:t},\bff_{1:t},\bfh)\sum_{i=1}^n w_i\delta(\bfh-\bfh_i)
\end{align*}
where we approximate $\phuman \approx \sum_{i=1}^n w_i\delta(\bfh-\bfh_i)$; we interpret $\bfh_i$ as \emph{waypoints, actuator inputs, or analyst insights}.  However, if the rank of the \emph{collective} human intent is $r$ and $k$ is the number of existing entries,~\cite{matrixcompletion} tells us that $n \geq Cn_1^{1.2}r \log N-k $ inputs can complete $n \to N$; in an important sense, we are completing a \emph{single} operator using the ``collective wisdom'' of the $n_1$ participants.  
\vspace{-0pt}

\textbf{Summary of Response:} \emph{IRT (lower bounding paradox) and optimal human collective representations demand a radical rethinking of coevolving ecosystems of humans, machines and adversaries: the distinction between individual and collective has been muddied in an unintuitive (yet mathematically precise) way.}
\vspace{-0pt}

\bibliographystyle{plainnat}
{\footnotesize
\bibliography{../standard-bibliography.bib}
}
\end{document}